\documentclass{article}

    \PassOptionsToPackage{numbers, compress}{natbib}

 \usepackage[preprint]{neurips_2025}

\usepackage[utf8]{inputenc} % allow utf-8 input
\usepackage[T1]{fontenc}    % use 8-bit T1 fonts
\usepackage{hyperref}       % hyperlinks
\usepackage{url}            % simple URL typesetting
\usepackage{booktabs}       % professional-quality tables
\usepackage{amsfonts}       % blackboard math symbols
\usepackage{nicefrac}       % compact symbols for 1/2, etc.
\usepackage{microtype}      % microtypography
\usepackage{xcolor}         % colors

\usepackage{amsmath,amssymb,amsfonts}
\usepackage{algorithmic}
\usepackage{graphicx} \graphicspath{ {figures/} } 
\usepackage{textcomp}
\usepackage{booktabs}
\usepackage{multirow}
\usepackage[rgb, dvipsnames, svgnames]{xcolor}
\usepackage{tcolorbox}
\tcbuselibrary{skins}
\tcbuselibrary{breakable}
\tcbuselibrary{listings}

\usepackage{tikz}
\usetikzlibrary{positioning, shapes, arrows.meta}
\usepackage[T1]{fontenc}
\usepackage{listings}
\usepackage{balance}
\usepackage{hyperref}

\title{CAP-IQA: Context-Aware Prompt-Guided CT Image Quality Assessment}

\author{%
  Kazi Ramisa Rifa\\
  \textit{ramisa.rifa@uky.edu} 
  \And
  Jie Zhang\\
  \textit{jie.zhang1@uky.edu}
  \And  
  Abdullah Imran\thanks{Corresponding author}\\
  \textit{aimran@uky.edu}
  \And
  \\[-8mm] University of Kentucky, Lexington, KY 40506, USA
}

\begin{document}

\maketitle

\begin{abstract}
  Prompt-based methods, which encode medical priors through descriptive text, have been only minimally explored for CT Image Quality Assessment (IQA).  While such prompts can embed prior knowledge about diagnostic quality, they often introduce bias by reflecting idealized definitions that may not hold under real-world degradations such as noise, motion artifacts, or scanner variability. To address this, we propose the Context-Aware Prompt-guided Image Quality Assessment (CAP-IQA) framework, which integrates text-level priors with instance-level context prompts and applies causal debiasing to separate idealized knowledge from factual, image-specific degradations. Our framework combines a CNN–based visual encoder with a domain-specific text encoder to assess diagnostic visibility, anatomical clarity, and noise perception in abdominal CT images. The model leverages radiology-style prompts and context-aware fusion to align semantic and perceptual representations. On the 2023 LDCTIQA challenge benchmark, CAP-IQA achieves an overall correlation score of 2.8590 (sum of PLCC, SROCC, and KROCC), surpassing the top-ranked leaderboard team (2.7427) by 4.24\%. Moreover, our comprehensive ablation experiments confirm that prompt-guided fusion and the simplified encoder-only design jointly enhance feature alignment and interpretability. Furthermore, evaluation on an in-house dataset of 91,514 pediatric CT images demonstrates the true generalizability of CAP-IQA in assessing perceptual fidelity in a different patient population.\\ 
  Our code is available at~\url{https://github.com/KaziRamisaRifa/capiqa}.
\end{abstract}

\section{Introduction}
\label{sec:introduction}
Computed tomography (CT) is indispensable in clinical diagnosis, although its quality may be compromised by noise, motion artifacts, and variations across scanners or acquisition protocols. Evaluating image quality automatically is important for reliable diagnosis and safer low-dose imaging. It enables scans with acceptable quality for diagnostic accuracy, helping to reduce radiation exposure without compromising clinical usefulness. Traditional objective metrics have been widely applied for CT image quality assessment (IQA), yet they generally emphasize low-level signal fidelity rather than clinical relevance~\cite{imran2021ssiqa}. This results in weak alignment with radiologists’ perceptual judgments, and the gap has motivated the shift toward no-reference (blind) IQA, where models directly predict perceptual or radiologist-like quality scores from CT images.

\begin{figure}
    \centering
    \includegraphics[width=0.63\linewidth]{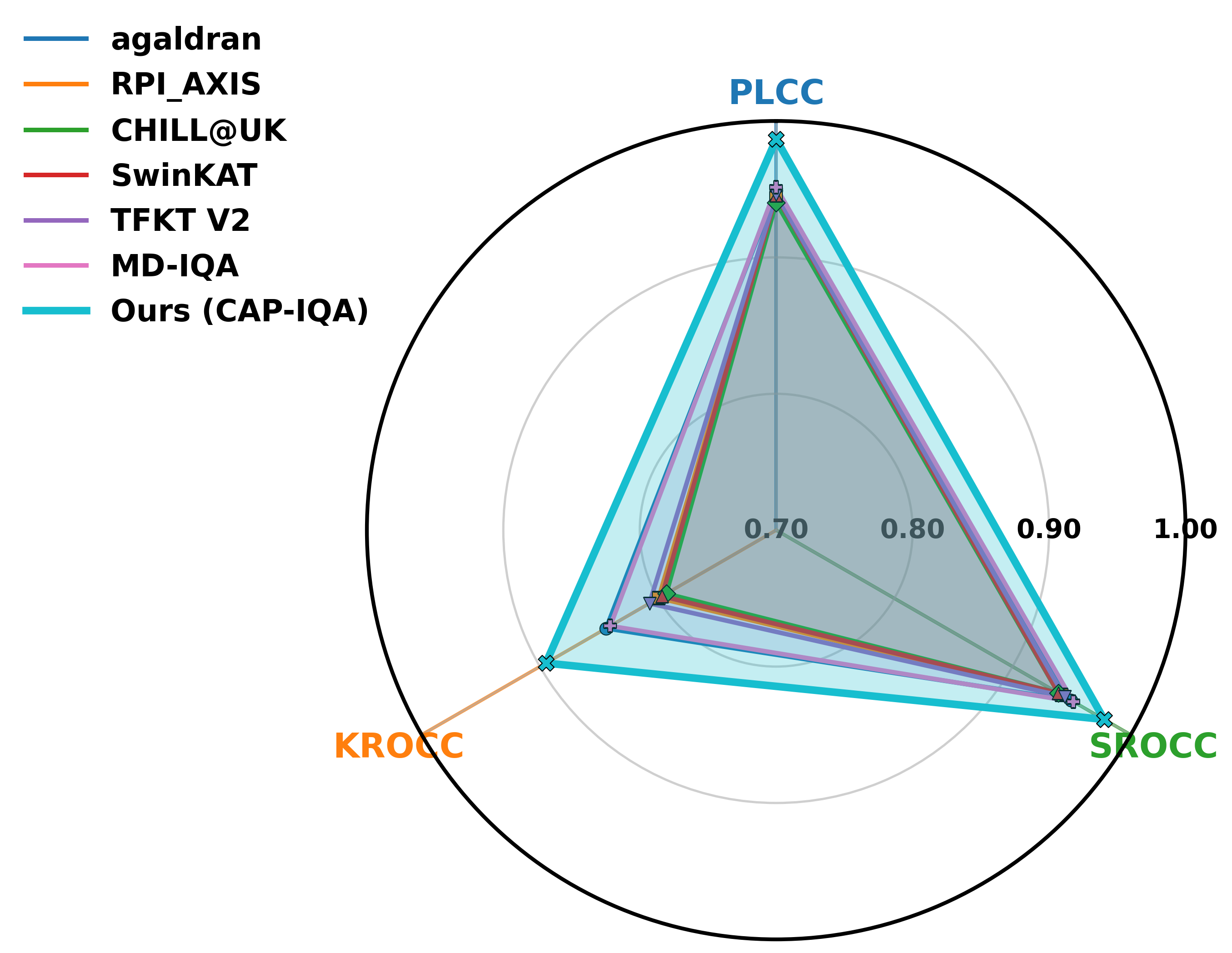}
    \caption{Our proposed CAP-IQA outperforms the top-ranked and recent models on the LDCTIQA 2023 benchmark~\citep{lee2025low}, achieving the highest correlation scores across Pearson’s linear correlation coefficient (PLCC), Spearman’s rank correlation coefficient (SROCC), and Kendall’s rank correlation coefficient (KROCC).}
    \label{fig:rader_plot}
\end{figure}

Early CT IQA relied heavily on simple noise measurements, which performed reasonably well for filtered back projection (FBP) reconstructions. However, the introduction of iterative reconstruction (IR) methods has made noise-based evaluation less reliable, as both the noise magnitude and texture vary with reconstruction algorithms. Conventional objective metrics, such as PSNR, SSIM, and MSE, have also been widely adopted; however, they primarily capture pixel-level discrepancies rather than diagnostic relevance~\cite{lee2025low}. Moreover, these methods typically require a clean reference image for comparison, which is an unrealistic expectation in clinical settings~\cite{shi2024blind}. As a result, their correlation with radiologists’ subjective quality judgments remains weak. 
Recent studies have explored task-specific IQA metrics focusing on clinical interpretability, including lesion detectability~\cite{gong2019deep}, anatomical visibility~\cite{lee2022no}, and diagnostic accuracy~\cite{lee2025low, mason2019comparison, zhang2025deep}. These approaches better reflect real-world diagnostic performance but are often difficult to generalize across various imaging conditions. Furthermore, their implementation in routine workflows is limited by dependence on specific clinical tasks and observer variability~\cite{imran2022personalized}.

Modern IQA developments span semi-supervised, transformer, and hybrid approaches. For example, a recent method~\cite{song2024md} employs multi-scale distribution regression with pseudo-labels to stabilize training on limited annotated data, despite being susceptible to different dose level noise and the heavy requirement for large unlabeled sets. Following prior work~\cite{yang2022maniqa}, a transformer-based model performs well on natural images; however, the 2D patch-based and computationally intensive design, which is not tailored to CT-specific artifacts, limits its potential for real-world clinical transfer.
The approach~\cite{lao2022attentions} leverages hybrid convolutional neural network (CNN)-transformer blocks with deformable convolution, although as a full-reference IQA method, it is inapplicable in CT settings where pristine reference scans do not exist. Most recently, Swin-KAT~\cite{rifa2025swin} integrates Kolmogorov–Arnold Networks~\cite{liu2024kan} with the Swin Transformer~\cite{liu2022swin} to deliver efficient reference-free CT IQA; nevertheless, it still relies more on labeled data, and its out-of-domain validation is limited in scope, leaving room for explicit generalization. Another line of work introduces generative priors; diffusion-based model~\cite{shi2024blind} employs the denoising diffusion probabilistic method to reconstruct a primary content image and compares it with distorted inputs using transformer evaluators. Despite accuracy gains, such methods are computationally expensive and not strictly non-reference, since they implicitly generate pseudo-references on which the IQA performance relies. Foundation models like the recent model~\cite{xun2025mediqa} attempt to unify IQA across modalities via large-scale pretraining and prompt-driven adaptation. While promising, the model depends on salient-slice heuristics and engineered prompt mappings, which may hinder flexibility under unseen CT distortions. However, most current IQA models often learn an overly general notion of good quality, which does not accurately reflect real CT degradations, such as noise, motion, or scanner differences. This can lead to poor generalization and mistakes in judging clinically usable images. To address the aforementioned issues, we propose a context-aware prompt-guided image quality assessment (CAP-IQA) framework that combines text priors with image-specific context prompts, thereby mitigating the bias of focusing solely on text prompts and leading to more reliable CT image quality assessment. 
Our contributions in the present paper can be summarized as follows:
\begin{itemize}
\item We introduce CAP-IQA, a framework for CT image quality assessment that explicitly mitigates the bias of focusing entirely on text prompts and produces more reliable predictions;
\item Unlike prior work that relies only on text prompts, CAP-IQA leverages both medical text priors and image-specific context prompts to better capture real CT degradations; 
\item Extensive evaluations on the public LDCTIQA challenge (LDCTIQAC) dataset demonstrate that CAP-IQA outperforms state-of-the-art CT IQA models by up to 4.24\% improvement in correlation with radiologists' assessments; 
\item Finally, effectiveness on our in-house pediatric CT dataset strongly suggests the capabilities of CAP-IQA in capturing robust and generalizable CT noise variations.
\end{itemize}

\begin{figure}
    \centering
    \includegraphics[width=\linewidth]{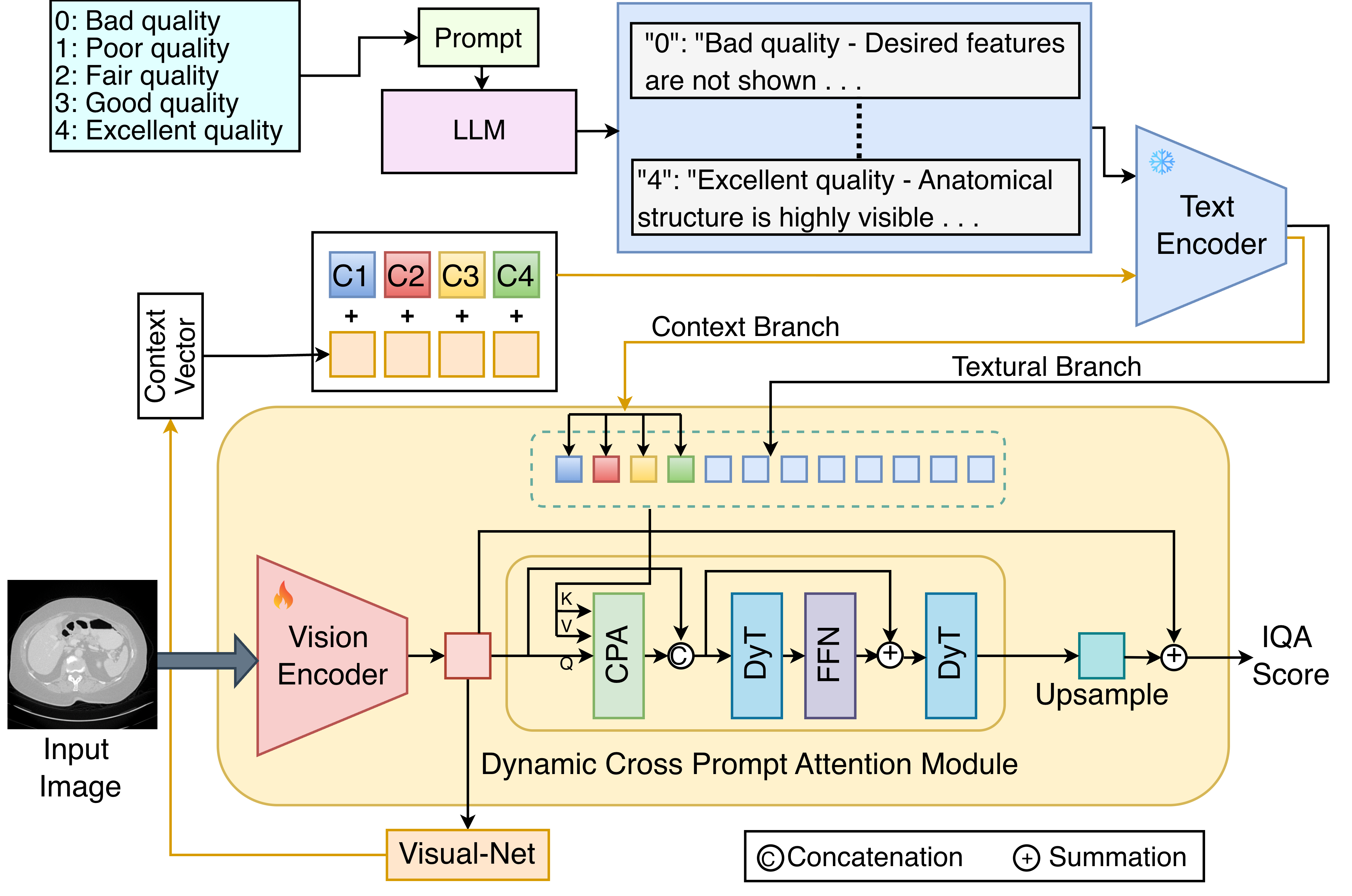}
    \caption{Overview of the proposed CAP framework. The textual branch encodes medical priors, and the context branch extracts image-specific prompts from CT images. Both are fused through cross-prompt attention, aligning medical knowledge with visual features to reduce non-relevant biases and improve CT image quality assessment.}
    \label{fig:method}
\end{figure}

\section{Related Work}

\subsection{Image Quality Assessment}
IQA has traditionally relied on handcrafted no-reference metrics~\cite{mittal2012no}, which are effective for natural images yet fail to capture CT-specific degradations. With the rise of deep learning, CNN-based and multi-task frameworks have been introduced to predict perceptual quality or detect artifacts directly from CT images. Reviews of CT IQA summarize hundreds of methods ranging from phantom-based evaluations for low-dose CT (LDCT) reconstruction, underscoring the increasing reliance on artificial intelligence (AI) in clinical imaging~\cite{xun2025charting, herath2025systematic}. 
Recent deep learning models process multiple CT views (e.g., axial, sagittal, and coronal slices together) and perceptual-quality benchmarks for LDCT have also been created~\cite{su2023deep}. Meanwhile, in MRI, similar efforts have led to datasets where radiologists have manually rated image quality~\cite{ma2024rad}. A novel direction is the combination of vision–language and generative models for IQA, where image captioning and large language models (LLMs) have been used to describe and score CT quality~\cite{chen2024iqagpt}. Despite these advances, existing IQA systems still lack semantic interpretability, as most focus solely on either text prompts or image features. While some approaches have used medical language models~\cite{yuan2022biobart}, they often lack context-awareness, as the text prompts may not be well-aligned with image features. This weak coupling between visual and textual representations restricts their ability to capture the complex relationships between image degradations and diagnostic quality in CT imaging.

\subsection{Vision-Language Models}

In recent years, vision–language models (VLMs) have shown increasing potential in medical imaging by aligning radiology images with textual knowledge. Contrastive pretraining on radiology images and reports (e.g., MedCLIP~\cite{wang2022medclip}, BioViL/BioViL-T~\cite{bannur2023learning}, BiomedCLIP~\cite{zhang2023biomedclip}) learns a shared embedding space that supports retrieval, classification, and report reasoning in medical images. Beyond contrastive encoders, instruction-tuned assistants such as LLaVA-Med~\cite{li2023llava} extend VLMs to conversational diagnosis and biomedical visual question answering (VQA) using PubMed Central (PMC) figure–caption corpora and GPT-generated instructions. Recent surveys synthesize this landscape and highlight emerging directions in clinical reasoning and multimodal report generation~\cite{lu2025integrating, ryu2025vision}, further include CT-based benchmarks for tumor analysis and large reviews summarizing VLM applications in 3D medical imaging~\cite{chen2025vision, wu2025vision, hartsock2024vision}. These works highlight the cross-modal representations but mostly focus on segmentation, classification, retrieval, report generation, or VQA, leaving IQA analysis underexplored. Early evidence from general-domain VLM-for-IQA (e.g., CLIP-based IQA) suggests that language priors can guide perceptual judgments, motivating medical adaptations~\cite{zhou2024uniqa, chen2024iqagpt}. The recent debiasing universal model~\cite{yun2025debiasing} shows that rich organ-level prompts from LLMs can inject helpful yet biased prior knowledge into vision models. To counter this, the authors add instance-level context prompts derived from the image itself and use a causal, debiasing training strategy that isolates the context contribution while suppressing prompt bias. Their framework (with a text branch, a context branch, and cross-prompt attention) improves multi-center generalization across several CT datasets, highlighting the value of context-aware prompts for robust medical image analysis. However, for image quality assessment (IQA), such context-aware and prompt-interactive mechanisms are largely unexplored, presenting an interesting direction for improving generalization in perceptual quality evaluation of CT images.

\subsection{Prompt-based Image Quality Assessment}
The emergence of prompt-based frameworks in IQA marks a shift from purely pixel-level metrics toward more adaptive, instruction-driven systems. For instance, PromptIQA~\cite{chen2024promptiqa} utilizes short sequences of image–score pairs as prompts, enabling the model to adapt to varied quality tasks without retraining. Multi‑Modal Prompt Learning IQA model~\cite{pan2024multi} introduces dual-text prompting (e.g., scene category + distortion type) in tandem with visual features to steer quality prediction. In the context of generated content, AI‑Generated IQA based on task‑specific prompt and multi‑granularity similarity (TSP-MGS)~\cite{xia2024ai} and PCQA~\cite{fang2024pcqa} leverage prompt-conditioning to evaluate prompt-conditioned image generation quality. Specifically for medical imaging, MedIQA~\cite{xun2025mediqa} presents a foundation model designed for multimodal medical IQA that aligns upstream imaging parameters with downstream prompt-annotated expert scores. Moreover, in the clinical context, IQAGPT~\cite{chen2024iqagpt} demonstrates that coupling a quality-captioning VLM with an LLM improves CT IQA over image-only baselines and CLIP-IQA variants, suggesting that clinical text can help make quality scoring more consistent and objective. These studies show that prompt-based IQA methods provide better adaptability and generalization, yet developing reliable and clinically consistent prompts while preserving the relationship between the image, its perceived quality, and the general prompt remains a significant challenge. To address this unresolved issue, our work follows this line by leveraging radiology-style prompts and diagnostic visibility to guide the model toward clinically meaningful features for enhancing robustness across diverse noise-aware CT image qualities.

\section{Method: CAP-IQA}

In this section, we introduce the proposed CAP-IQA framework, a prompt-driven image quality assessment (IQA) model designed for 2D CT images. CAP-IQA combines contextual prompts with medical text priors for predicting perceptual quality scores in CT images. It explicitly integrates medical knowledge prompts with adaptive and instance-level context prompts (Fig.~\ref{fig:method}).

Let $I \in \mathbb{R}^{1 \times H \times W}$ denote an input CT image, and $y \in \mathbb{R}$ denote the expert-provided quality score. The dataset is denoted as $\mathcal{D} = \{(I_i, y_i)\}_{i=1}^N$. The goal is to learn a mapping $f: I \mapsto \hat{y}$ such that $\hat{y}$ is highly correlated with radiologist assessments.

\subsection{Textual Branch: Medical Priors}
We leverage an LLM to obtain descriptive medical prompts $T$ for each image, capturing anatomical visibility, artifact presence, and noise characteristics. These prompts are embedded via a frozen domain-specific language model (PubMedBERT~\cite{gu2021domain}):
\begin{equation}
t = \Phi(T), \qquad t \in \mathbb{R}^{d},
\end{equation}
where $d$ (e.g., 768) is the text embedding dimension. These priors encode high-level, knowledge-rich features; however, relying solely on text-based priors can introduce limitations, such as prompts may emphasize standardized definitions of quality while failing to capture subtle, patient-specific or scan-dependent degradations that fall outside typical diagnostic descriptions. As a result, global textual knowledge may overlook rare artifacts, atypical anatomical presentations, or context-dependent noise patterns that are not explicitly described in medical guidelines. To address this, our framework is designed to complement universal text priors with instance-level context prompts. This ensures that both high-level semantic information and individual scan characteristics contribute to the final quality assessment.

\subsection{Visual Encoder and Context Branch}
The input image $I\in \mathbb{R}^{1 \times H \times W}$ is processed by a U-Net babsed encoder~\cite{ronneberger2015u} to produce a bottleneck feature map $z \in \mathbb{R}^{C \times H' \times W'}$, where $C$ is the number of channels and $H', W$ are the spatial dimensions of the deepest encoder layer. For each spatial position $(h, w)$, the notation $z[:, h, w]$ refers to the vector of all $C$ channels at that location. That is, $z[:, h, w] \in \mathbb{R}^C$ contains the feature values for every channel at the position $(h, w)$ in the feature map. To obtain image feature embedding, we apply global average pooling: 
\begin{equation}
    f = \frac{1}{H'W'} \sum_{h=1}^{H'} \sum_{w=1}^{W'} z[:, h, w], \qquad f \in \mathbb{R}^C,
\end{equation}
where $f$ is the pooled feature vector for the CT image.

To capture image-specific information and mitigate the limitations of static prior knowledge, we introduce a set of $L$ learnable context tokens $\{c_1, ..., c_L\}$, with each $c_l \in \mathbb{R}^d$ ($d$ is the prompt dimension, typically $768$). These are adaptively updated for each image by projecting the pooled feature $f$ via a lightweight multilayer perceptron (MLP), resulting in image-conditioned tokens:
\begin{equation}
    c'_l = \mathrm{MLP}(f) + c_l, \qquad l = 1, ..., L,
\end{equation}
where $\mathrm{MLP}: \mathbb{R}^C \rightarrow \mathbb{R}^d$. The final prompt set is formed by concatenating the context tokens with a fixed text prompt $t \in \mathbb{R}^{1 \times d}$, which encodes global, IQA-level prior knowledge (from a pretrained language model):
\begin{equation}
    \pi = \begin{bmatrix} t \\ c'_1 \\ \vdots \\ c'_L \end{bmatrix} \in \mathbb{R}^{(L+1) \times d}.
\end{equation}

Instance-adaptive context prompts enable the model to adjust its features based on each scan, rather than relying only on general medical knowledge. This helps reduce bias from using fixed text prompts and makes the model more responsive to unique or unusual cases. By combining general information and scan-specific context, the model can better handle images that are different from the training data and produce more reliable image assessments for rare or challenging cases. This approach can lead to more fair and accurate IQA score predictions.

\subsection{Dynamic Cross-Prompt Attention (DCPA)}

To combine the pooled image feature and the prompt information, we use a Dynamic Cross-Prompt Attention (DCPA) mechanism. The pooled visual feature $f \in \mathbb{R}^C$ is first projected to the prompt embedding dimension $d$ using a learned weight matrix $W_q \in \mathbb{R}^{d \times C}$, and then normalized:
\begin{equation}
    q = \mathrm{LayerNorm}(W_q f), \quad q \in \mathbb{R}^{1 \times d}
\end{equation}
The prompt matrix $\pi \in \mathbb{R}^{(L+1) \times d}$, which is built by stacking the text prompt and $L$ context prompts, is also normalized to produce the key and value matrices:
\begin{equation}
    k = v = \mathrm{LayerNorm}(\pi), \quad k, v \in \mathbb{R}^{(L+1) \times d}
\end{equation}
Using these, we compute multi-head attention, where the query $q$ attends to the keys $k$ and values $v$ to produce an attended feature $a \in \mathbb{R}^{1 \times d}$:
\begin{equation}
    a = \mathrm{MultiHeadAttn}(q, k, v).
\end{equation}
Next, we concatenate the attended feature $a$ with the original query $q$ along the feature axis, giving $x = [a; q] \in \mathbb{R}^{1 \times 2d}$. Unlike many existing architectures that use LayerNorm after attention, we use a Dynamic Tanh (DyT) operation~\cite{zhu2025transformers}. Specifically, DyT is defined as
\begin{equation}
    \mathrm{DyT}(x) = \tanh(\alpha x) \odot w + b,
\end{equation}
where $\alpha \in \mathbb{R}^{2d}$ is a learnable scaling vector, $w, b \in \mathbb{R}^{2d}$ are per-channel scale and bias parameters, and $\odot$ denotes elementwise multiplication. After DyT, $x$ is passed through a two-layer feed-forward network (FFN) with a ReLU activation, and then added back to the input via a residual connection. The result is again processed with DyT to form the final DCPA output. Finally, we project this output back to the encoder feature dimension with a learned matrix $W_p \in \mathbb{R}^{C \times 2d}$, yielding $c_f = W_p c \in \mathbb{R}^{1 \times C}$.

The use of DyT in place of standard normalization techniques, such as LayerNorm, brings several benefits. Unlike LayerNorm, which depends on computing mean and variance over features and can be sensitive to the choice of batch size or distribution shift, DyT applies a simple non-linear transformation to each feature independently. This approach helps keep activations within a stable range, reduces training instability caused by outliers, and removes the reliance on batch statistics. Recent work has shown that models using DyT maintain strong performance while being more robust and efficient, particularly in settings with small batch sizes or highly variable data distributions. In our context, the use of DyT within the DCPA block allows the model to reliably fuse visual and prompt information while ensuring stability and flexibility for diverse and challenging medical CT images.

\subsection{Feature Fusion and Regression}

After obtaining the DCPA-derived feature vector $c_f \in \mathbb{R}^{1 \times C}$, we broadcast it across the spatial dimensions and add it to the encoder output feature map $z \in \mathbb{R}^{C \times H' \times W'}$. This results in a fused representation: $\tilde{z} = z + \text{Expand}(c_f)$. The fused feature map $\tilde{z}$ is then processed by a regression head to generate the final IQA score. First, we apply adaptive average pooling: $h = \text{AdaptiveAvgPool2d}(\tilde{z}) \in \mathbb{R}^{C \times 1 \times 1}$. Next, we flatten $h$ into a vector $h' \in \mathbb{R}^C$, and compute the predicted score as $\hat{y} = \sigma(W_r h')$, where $W_r$ is a linear layer and $\sigma$ denotes the sigmoid activation. The predicted score is scaled to $[0,4]$ to align with the LDCTIQA scoring criteria (Table~\ref{tab:score-criteria}).

\subsection{Loss Function}
The model is trained using mean squared error (MSE) between the predicted and ground-truth scores:
\begin{equation}
    \mathcal{L} = \frac{1}{N} \sum_{i=1}^N \left( \hat{y}_i - y_i \right)^2,
\end{equation}
where $\hat{y}_i$ is the predicted score, $y_i$ is the ground-truth score, and $N$ is the batch size.

\section{Experiments and Results}
\subsection{Data} 
We utilize the 2023 LDCTIQA challenge dataset~\cite{lee2025low}. The dataset consists of abdominal low-dose CT images that are annotated with perceptual quality scores ranging from 0 (worst quality) to 4 (best quality), obtained by averaging ratings from multiple experienced radiologists (Table~\ref{tab:score-criteria}). The LDCTIQA dataset consists of a total of 1,300 public abdominal CT images from 15 patient scans, including 10 from the Mayo Clinic in the United States and 5 from the National Cancer Center in South Korea, ensuring diversity across institutions. Among these, 1,000 images were kept for training, generated from 50 slices of seven patients, while 300 images generated from 30 slices of five patients were used for testing. The dataset was constructed using a physics-based pipeline that introduced realistic noise and streak artifacts across different dose levels, and each image was rated by radiologists for diagnostic quality~\cite{lee2025low}, making it the first large-scale open-access benchmark for low-dose CT image quality assessment. For our experiments, we follow the data settings of TFKT V2~\cite{rifa2025tfkt}. A subset (100 images) of the training data was used as the validation set.

\begin{table}
\caption{Image scoring criteria for the 2023 low-dose CT image quality assessment (LDCTIQA) Grand Challenge.}
\label{tab:score-criteria}
\centering
% \resizebox{\linewidth}{!}{
\begin{tabular}{cc lc l}
\toprule
Score && Quality && Diagnostic quality criteria \\
\midrule
0 && Bad && Desired features are not shown\\
1 && Poor && Diagnostic interpretation is impossible\\
2 && Fair && Suitable for compromised interpretation\\
3 && Good && Good for diagnostic interpretation\\
4 && Excellent && Anatomical features are clearly visible\\
\bottomrule
\end{tabular}
    % }
\end{table}

\subsection{Implementation Details} The proposed model was trained for 100 epochs with a minibatch size of 8.  We implemented the model in Python with the PyTorch framework. The model is run on an Intel(R) Xeon(R) W7-2475X processor (2600MHz) with 128 GB RAM and dual NVIDIA A4000X2 GPUs (32GB). We utilized the AdamW optimizer with a warm-up (10 epochs) cosine scheduler, setting the initial learning rate to $1 \times 10^{-4}$. 

\subsection{Evaluation Measures}
Following~\cite{lee2025low, rifa2025tfkt}, we assess the model’s performance using both linear and non-linear correlation coefficients to ensure a comprehensive evaluation. Specifically, we report Pearson’s linear correlation coefficient $(r)$, Spearman’s rank correlation coefficient $(\rho)$, and Kendall’s rank correlation coefficient $(\tau)$. Given the reference and predicted IQA scores, these three correlation coefficients are computed as follows:

\begin{equation}
\label{eq:r}
r = \frac{\sum_{i=1}^n(y_i - \bar{y})(\hat{y}_i - \bar{\hat{y}})}{\sum{i=1}^n(y_i-\bar{y})^2\sum_{i=1}^n(\hat{y}_i-\bar{\hat{y}})^2},
\end{equation}
where $\bar{y}$ and $\bar{\hat{y}}$ represent the mean values of the reference scores $y$ and the predicted scores $\hat{y}$, respectively.

\begin{equation}
\label{eq:rho}
\rho = 1 - \frac{6\sum^n_{i=1}d_i^2}{n(n^2 - 1)},
\end{equation}
where $d_i = \hat{y}_i - y_i$ denotes the rank difference between the $i$-th image in the reference and predicted assessments.

\begin{equation}
\label{eq:tau}
\tau = \frac{P - Q}{\sqrt{(P + Q + y_0)(P + Q + \hat{y}_0)}},
\end{equation}
where $P$ and $Q$ denote the numbers of concordant and discordant pairs, respectively; $y_0$ is the number of pairs tied only in the reference scores, and $\hat{y}_0$ is the number of pairs tied only in the predicted scores. Finally, the overall performance is quantified by aggregating the values of all three correlation coefficients. 

\begin{equation}
\label{eq:s}
s = |r| + |\rho| + |\tau|.
\end{equation}

\subsection{Results and Discussion} 
The fusion mechanism in CAP-IQA yields a measurable improvement in correlation metrics, underscoring the benefit of jointly modeling visual and textual understanding. The model achieves higher stability and better agreement with radiologist scores compared to its vision-only prior models, which indicates the text-guided representation enhances the perception of diagnostic quality rather than pixel noise. Moreover, the results suggest that the proposed fusion operation within the dynamic cross-prompt attention module helps the model to interpret image quality more contextually and align better with clinical perception.

\begin{table}
\centering
\caption{Quantitative comparison among state-of-the-art methods and CAP-IQA, evaluated on the validation set. We report the individual correlation coefficients and the overall coefficient scores.}
\label{tab:results}
\resizebox{0.63\linewidth}{!}{
\begin{tabular}{lcccc}
\toprule
\textbf{Method} & \textbf{$r$} & \textbf{$\rho$} & \textbf{$\tau$} & \textbf{$s$} \\
\midrule
MD-IQA \cite{song2024md} & 0.9771 & 0.9793 & 0.9106 & 2.8670\\
DBCNN \cite{zhang2018blind} & 0.9714 & 0.9734 & 0.8808 & 2.8255\\
MANIQA \cite{yang2022maniqa} & 0.9768 & 0.9786 & 0.8891 & 2.8445\\
QPT \cite{zhao2023quality} & 0.9743 & 0.9732 & 0.8797 & 2.8272\\
AHIQ \cite{lao2022attentions} & 0.9762 & 0.9746 & 0.8810 & 2.8317\\
TReS \cite{golestaneh2022no} & 0.9755 & 0.9745 & 0.8786 & 2.8286\\
SSIQA \cite{imran2021ssiqa} & 0.9784 & 0.9767 & 0.8905 & 2.8456\\
Swin-KAT~\cite{rifa2025swin}  & 0.9831 & 0.9825 & 0.9031 & 2.8687\\
D-BIQA \cite{shi2024blind} & 0.9814 & 0.9816 & 0.9122 & 2.8753\\
TFKT V2~\cite{rifa2025tfkt} & 0.9858 & 0.9861 & 0.9171 & 2.8890\\
\textbf{CAP-IQA (Ours)} & \textbf{0.9864} & \textbf{0.9870} & \textbf{0.9187} & \textbf{2.8921} \\
\bottomrule
\end{tabular}
}
\end{table}

Table~\ref{tab:results} reports the quantitative performance comparison among CAP-IQA and several state-of-the-art IQA methods on the validation set. The results demonstrate that CAP-IQA consistently outperforms all existing approaches across all correlation metrics. Compared to the baseline TFKT V2~\cite{rifa2025swin}, which is a hybrid CNN-Transformer architecture, CAP-IQA achieves a 0.06\% improvement in $r$, 0.09\% in $\rho$, and 0.17\% in $\tau$. In contrast, our model maintains strong and stable generalizability across both validation and test sets. These results indicate that the proposed context prompt mechanism, integrated with the CNN-based encoder and transformer-like architecture, provides superior performance compared to state-of-the-art architectures. The prompt-guided feature interaction increases the contextual understanding, resulting in more accurate quality predictions for low-dose CT images. CAP-IQA achieves higher overall performance than other strong baselines, including D-BIQA~\cite{shi2024blind} and Swin-KAT~\cite{rifa2025swin}, with relative improvements of 0.58\% and 0.82\%, respectively. These consistent improvements demonstrate that the proposed context-aware prompting enables the model to align noise features better, leading to more accurate and reliable quality predictions for low-dose CT images.

The radar plot in Fig.~\ref{fig:rader_plot} compares the correlation metrics of CAP-IQA with top-performing LDCTIQA challenge and recent CT IQA models. The plot illustrates that CAP-IQA consistently forms the outermost boundary across all axes, indicating stronger agreement with radiologist scores compared to all other IQA models. Moreover, Table~\ref{tab:ldctiqac_comparison} presents a comprehensive comparison of CAP-IQA with the top-performing teams from the LDCTIQA challenge as well as several recent CT-specific and hybrid IQA models. The proposed CAP-IQA achieves the highest overall score of 2.8590, outperforming all prior methods across all correlation metrics ($r$ = 0.9866, $\rho$ = 0.9775, $\tau$ = 0.8949). Compared with the strongest leaderboard team, agaldran (2.7427), CAP-IQA delivers an absolute gain of 0.1163, reflecting a 4.24\% improvement in overall correlation. Even when compared with recent advanced frameworks such as MD-IQA (2.7431) and Med-IQA (0.9764 for $r$), CAP-IQA shows more consistent alignment with both radiologist scores and model-based evaluations. We performed the Wilcoxon signed-rank test by taking the error in predictions for the test data of each of the TFKT V2 and CAP-IQA models. The test confirmed that our CAP-IQA is significantly better than TFKT V2~($p-value <$ 0.001). While TFKT V2 and some other models performed well on the validation set, there are big drops in their test performances. This can be attributed to the fact that the test IQA references were obtained by averaging scores assigned by six radiologists. However, our proposed CAP-IQA model demonstrates a strong generalization capability driven by its context-aware prompt learning.

\begin{table}
    \centering
    \caption{Quantitative comparison among the top-ranked teams in LDCTIQAC 2023 \cite{lee2025low} and CAP-IQA, evaluated on the LDCTIQAC 2023 test set.}
    \label{tab:ldctiqac_comparison}
    \resizebox{0.83\linewidth}{!}{
    \begin{tabular}{l l cccc}
    \toprule
    \textbf{Team} & \textbf{Model} & \textbf{$r$} & \textbf{$\rho$} & \textbf{$\tau$} & \textbf{$s$} \\
    \midrule
    agaldran~\cite{lee2025low}     & Swin \& BiTResNeXt50 & 0.9491 & 0.9495 & 0.8440 & 2.7427 \\
    RPI\_AXIS~\cite{lee2025low}    & MANIQA             & 0.9434 & 0.9414 & 0.7995 & 2.6843 \\
    CHILL@UK~\cite{lee2025low}     & EfficientNet-V2L   & 0.9402 & 0.9387 & 0.7930 & 2.6719 \\
    FeatureNet~\cite{lee2025low}   & ViT\&GLCM          & 0.9362 & 0.9338 & 0.7851 & 2.6550 \\
    Team Epoch~\cite{lee2025low}   & EDCNN              & 0.9278 & 0.9232 & 0.7691 & 2.6202 \\
    gabybaldeon~\cite{lee2025low}  & CNN-ViT            & 0.9143 & 0.9096 & 0.7432 & 2.5671 \\
    \midrule
    SwinKAT~\cite{rifa2025swin}      & MLP-KAN Transformer & 0.9454 & 0.9389 & 0.7967 & 2.6811 \\
    TFKT V2~\cite{rifa2025tfkt}      & Hybrid CNN-Transformer     & 0.9462 & 0.9440 & 0.8073 & 2.6975 \\
    MD-IQA~\cite{song2024md}       & Multi-Scale Distribution   & 0.9513 & 0.9514 & 0.8404 & 2.7431 \\
    Med-IQA~\cite{xun2025mediqa}      & MANIQA Prompt-Based        & 0.9764 & 0.9762 & --      & -- \\
    Ours         & Context-Aware Prompt-Based & \textbf{0.9866} & \textbf{0.9775} & \textbf{0.8949} & \textbf{2.8590} \\
    \bottomrule
    \end{tabular}
    }
\end{table}

\begin{figure}
\centering
\resizebox{0.85\linewidth}{!}{
\begin{tabular}{c c c}

\includegraphics[width=0.33\linewidth]{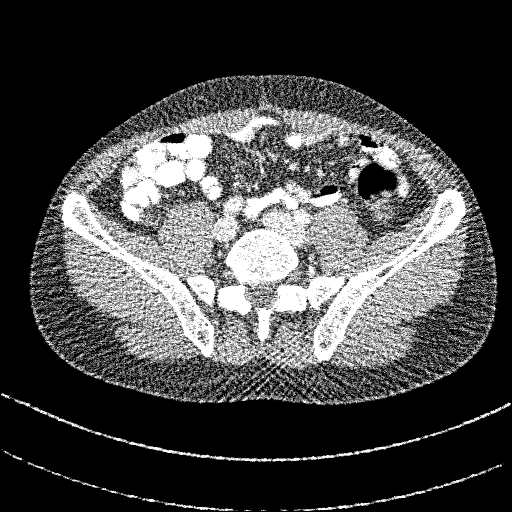}
&
\includegraphics[width=0.33\linewidth]{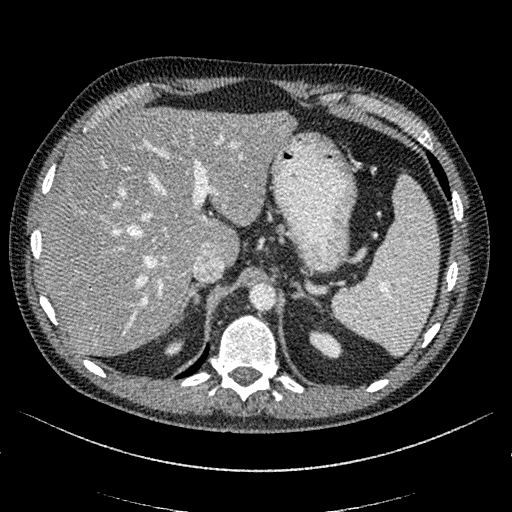}
&
\includegraphics[width=0.33\linewidth, trim={0in 0in 0in 0in}, clip]{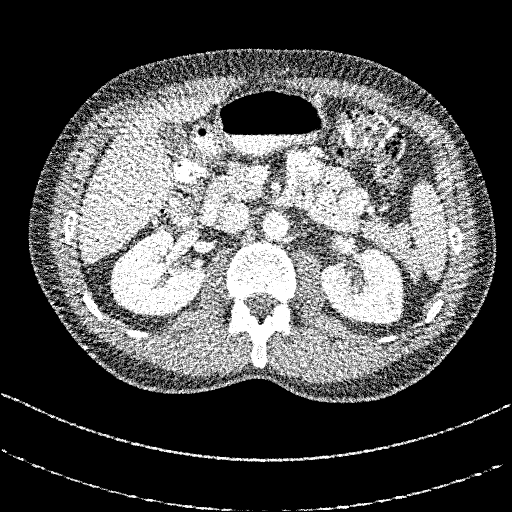}\\
{\large 0.31 [0.33]} & {\large 2.94 [2.83]} & {\large 0.12 [0.00]}
\smallskip\\
\includegraphics[width=0.33\linewidth]{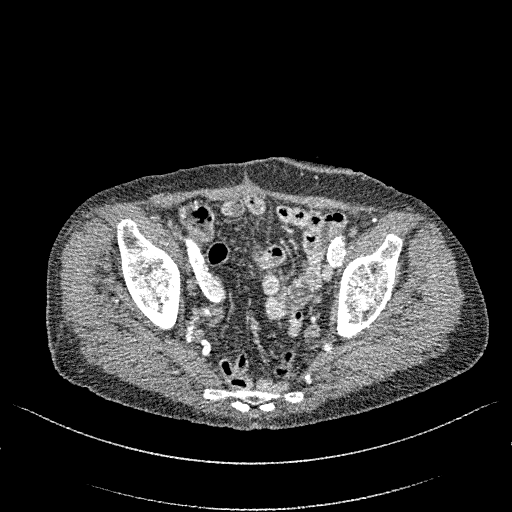}
&
\includegraphics[width=0.33\linewidth, trim={0in 0in 0in 0in}, clip]{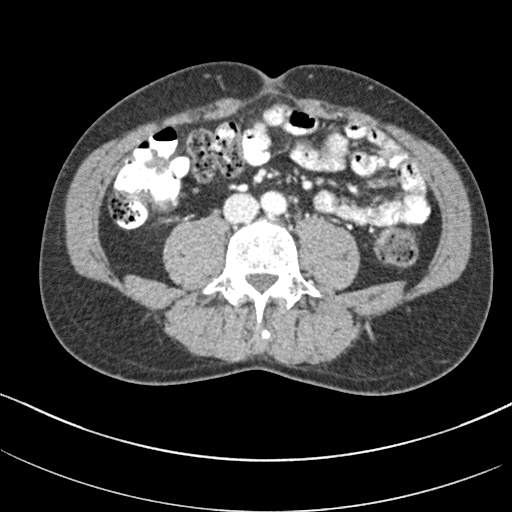}
&
\includegraphics[width=0.33\linewidth, trim={0in 0in 0in 0in}, clip]{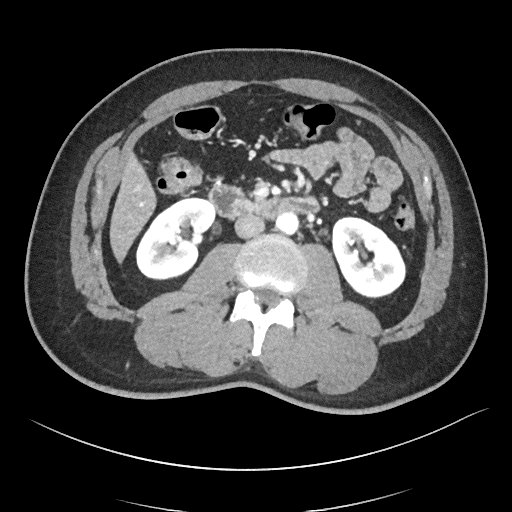}\\
{\large 2.02 [2.00]} & {\large 3.82 [3.83]} & {\large 3.85 [4.00]} 
\end{tabular}
}
\caption{Effectiveness of our CAP-IQA model in accurately assessing the quality of abdominal CT images across diverse IQA scores. Model predictions are in good agreement with the [actual] scores. CT images are visualized after removing MATLAB’s +1024 HU offset and re-windowing to Window Width: 400 and Window Level: 50.}
\label{fig:iqa_comparison}
\end{figure}

\subsection{Ablation Studies} 
An ablation study was conducted to investigate the effect of different feature fusion strategies in CAP-IQA, as shown in Table~\ref{tab:opp_ablation}. The model is experimented with Sum and Concat operations applied after the cross-prompt attention module (CPA), feed-forward network (FFN), and upsampling layers. The configuration CPA-Out: Concat, FFN-Out: Sum, Up-Out: Sum achieved the best overall score of 2.8556, which indicates that concatenation at the CPA stage effectively enriches feature representation while maintaining stability in later layers. In contrast, applying concatenation at all stages resulted in slight performance degradation due to redundant feature aggregation. These results suggest that a selective combination of concatenation and summation operations provides the most balanced and efficient fusion for CT image quality prediction. 

Furthermore, we compared the performance of CAP-IQA using standard LayerNorm and Dynamic Tanh (DyT) normalization. The DyT variant achieves a slightly higher overall score (2.8556 vs. 2.8549) and marginally better correlation metrics, suggesting that DyT provides smoother feature scaling and better stability during training. Unlike LayerNorm, which normalizes features linearly, DyT introduces a nonlinear scaling using the tanh function, allowing smoother feature transitions and preventing activation saturation. This dynamic adjustment helps preserve subtle intensity variations that are critical for CT quality assessment. As a result, DyT enhances the model’s ability to capture fine-grained contrast and noise patterns, which leads to more consistent and perceptually accurate IQA predictions.

\begin{table}[t]
    \centering
    \caption{Ablation study of different strategies in CAP-IQA where \textbf{CPA-Out}, \textbf{FFN-Out}, and \textbf{Up-Out} denote the feature fusion operations applied after the cross-prompt attention (CPA), feed-forward network (FFN), and upsampling layers, respectively.}
    % \vspace{0.2cm}
    \label{tab:opp_ablation}
    \resizebox{\linewidth}{!}{
    \begin{tabular}{l cccc}
    \toprule
    \textbf{Operation Configuration} & \textbf{$r$} & \textbf{$\rho$} & \textbf{$\tau$} & \textbf{$s$} \\
    \midrule
    CPA-Out: Sum, FFN-Out: Sum, Up-Out: Sum 
        & 0.9830 $\pm$ 0.0010 & 0.9796 $\pm$ 0.0013 & 0.8911 $\pm$ 0.0041 & 2.8537 $\pm$ 0.0067 \\
    \midrule
    CPA-Out: Concat, FFN-Out: Sum, Up-Out: Sum 
        & \textbf{0.9836 $\pm$ 0.0009} & \textbf{0.9805 $\pm$ 0.0011} & \textbf{0.8915 $\pm$ 0.0041} & \textbf{2.8556 $\pm$ 0.0057} \\
    \midrule
    CPA-Out: Concat, FFN-Out: Concat, Up-Out: Sum 
        & 0.9821 $\pm$ 0.0015 & 0.9798 $\pm$ 0.0012 & 0.8905 $\pm$ 0.0046 & 2.8524 $\pm$ 0.0071 \\
    \midrule
    CPA-Out: Concat, FFN-Out: Concat, Up-Out: Concat 
        & 0.9796 $\pm$ 0.0018 & 0.9771 $\pm$ 0.0015 & 0.8897 $\pm$ 0.0059 & 2.8464 $\pm$ 0.0092 \\
    \midrule
    CPA-Out: Concat, FFN-Out: Sum, Up-Out: Concat 
        & 0.9812 $\pm$ 0.0016 & 0.9781 $\pm$ 0.0013 & 0.8886 $\pm$ 0.0058 & 2.8479 $\pm$ 0.0087 \\
    \bottomrule
    \end{tabular}
    }
\end{table}

\begin{table}[t]
\centering
\caption{Performance comparison of various vision and text encoder architectures integrated within the proposed CAP-IQA model, evaluated on the LDCTIQAC 2023 test dataset.}
\label{tab:vision_comp}
\resizebox{\linewidth}{!}{
\begin{tabular}{l l cccc}
\toprule
\textbf{Vision Encoder} & \textbf{Text Encoder} & \textbf{$r$} & \textbf{$\rho$} & \textbf{$\tau$} & \textbf{$s$} \\
\midrule
EfficientNetV2L~\cite{tan2021efficientnetv2} & \multirow{6}{*}{\centering PubMedBert~\cite{gu2021domain}} &
0.9730 $\pm$ 0.0012 & 0.9624 $\pm$ 0.0043 & 0.8380 $\pm$ 0.0114 & 2.7734 $\pm$ 0.0165 \\
Vision Transformer~\cite{dosovitskiy2020image} & &
0.9747 $\pm$ 0.0016 & 0.9756 $\pm$ 0.0005 & 0.8752 $\pm$ 0.0011 & 2.8257 $\pm$ 0.0010 \\
Swin Transformer V2~\cite{liu2022swin} & &
0.9721 $\pm$ 0.0063 & 0.9513 $\pm$ 0.0042 & 0.8121 $\pm$ 0.0101 & 2.7355 $\pm$ 0.0056 \\
BiomedCLIP~\cite{zhang2023biomedclip} & &
0.9638 $\pm$ 0.0053 & 0.9595 $\pm$ 0.0032 & 0.8332 $\pm$ 0.0077 & 2.7564 $\pm$ 0.0159 \\
MedVAE~\cite{varma2025medvae} & &
\underline{0.9808 $\pm$ 0.0013} & \underline{0.9758 $\pm$ 0.0004} & \underline{0.8774 $\pm$ 0.0015} & \underline{2.8339 $\pm$ 0.0009} \\
CNN-Encoder & &
\textbf{0.9836 $\pm$ 0.0009} & \textbf{0.9805 $\pm$ 0.0011} & \textbf{0.8915 $\pm$ 0.0041} & \textbf{2.8556 $\pm$ 0.0057} \\
\midrule
BiomedCLIP~\cite{zhang2023biomedclip} & \multirow{3}{*}{\centering BiomedCLIP~\cite{zhang2023biomedclip}} &
0.9753 $\pm$ 0.0010 & 0.9616 $\pm$ 0.0020 & 0.8321 $\pm$ 0.0074 & 2.7690 $\pm$ 0.0102 \\
MedVAE~\cite{varma2025medvae} & &
0.9821 $\pm$ 0.0012 & 0.9730 $\pm$ 0.0017 & 0.8770 $\pm$ 0.0020 & 2.8321 $\pm$ 0.0024 \\
CNN-Encoder &  &
0.9723 $\pm$ 0.0014 & 0.9597 $\pm$ 0.0032 & 0.8300 $\pm$ 0.0088 & 2.7620 $\pm$ 0.0131 \\
\bottomrule
\end{tabular}
}
\end{table}

Table~\ref{tab:vision_comp} compares different vision encoders paired with medical text encoders for CT image quality prediction. The configuration using a CNN-Encoder as the vision backbone with PubMedBERT as the text encoder achieves the best overall performance (2.8556), outperforming transformer-based models such as MedVAE (2.8339) and Vision Transformer (2.8257). Although transformer backbones capture long-range dependencies, the CNN-based encoder provides stronger spatial and quality representations, which are more effective for assessing CT image quality. These results suggest that convolutional features, when guided by medical text priors, offer a more stable and reliable prediction of CT IQA.

\subsection{Range-wise Performance Analysis}
\begin{figure}[t]
    \centering
    \includegraphics[width=0.85\linewidth]{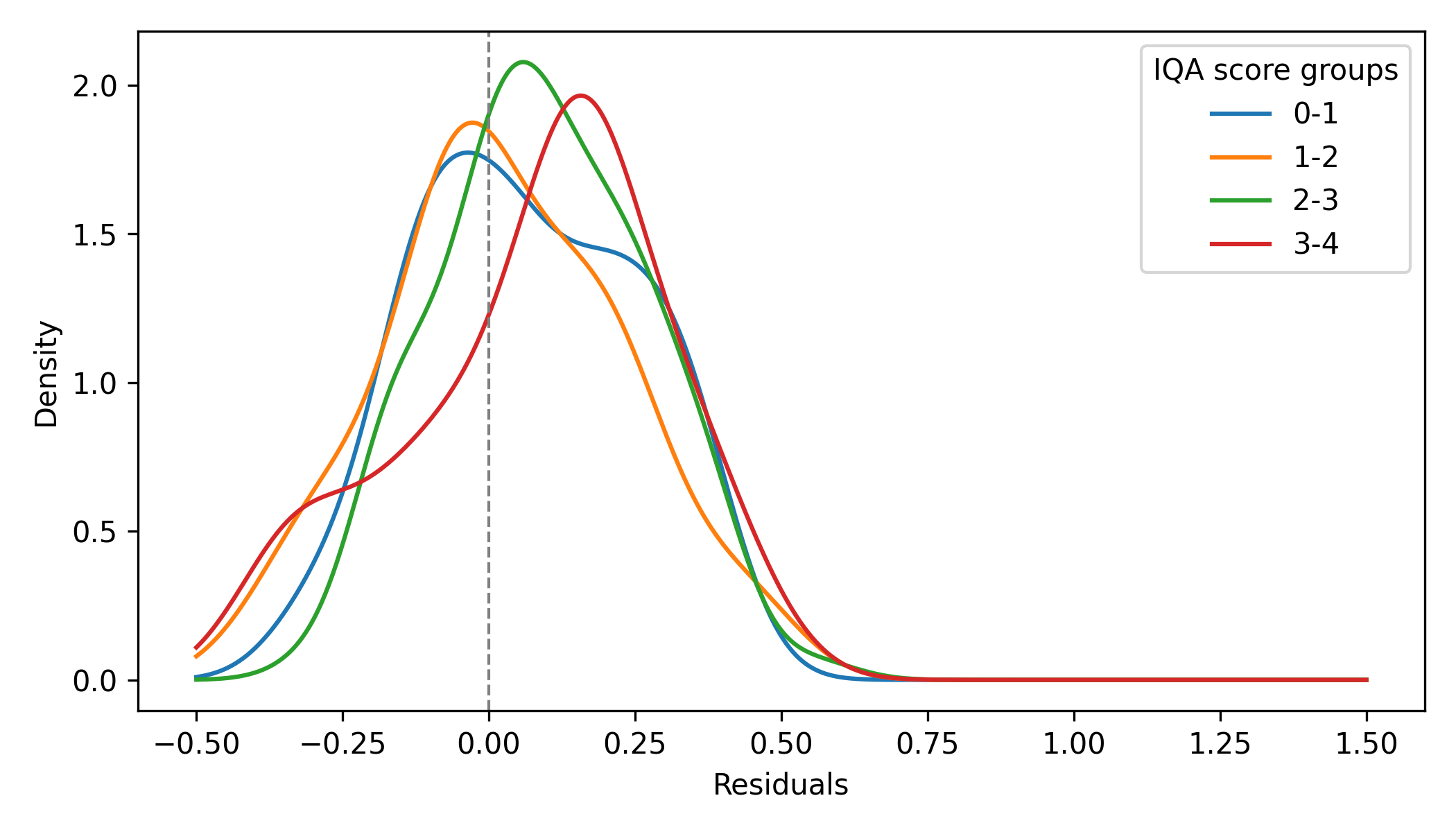}
    \caption{Kernel density plot across IQA score groups. Predictions remain centered near zero across all groups, indicating minimal bias of the model across different image quality levels.}
    \label{fig:kde_iqa_groups}
\end{figure}

Fig.~\ref{fig:iqa_comparison} visualizes sample abdominal CT images with varying quality levels from the test set. The predicted IQA scores show strong agreement with the reference scores across different noise and artifact intensities, which indicates the robustness of CAP-IQA in accurately quantifying CT image quality. The kernel density plot in Fig.~\ref{fig:kde_iqa_groups} illustrates how prediction residuals are distributed across different IQA score groups. All curves are centered near zero, indicating balanced predictions without systematic bias toward any specific quality range. Moreover, the box plot presents the distribution of absolute prediction errors across different IQA score groups. The median errors remain low and consistent, showing that CAP-IQA performs reliably across all quality levels, with minimal variations and stability in predictions. Consistent with this interpretation, our pairwise testing using the Kruskal-Wallis H-test further confirms that there is no significant difference in performance for different IQA score groups (p-value $\approx$ 0.38).

\begin{figure}[t]
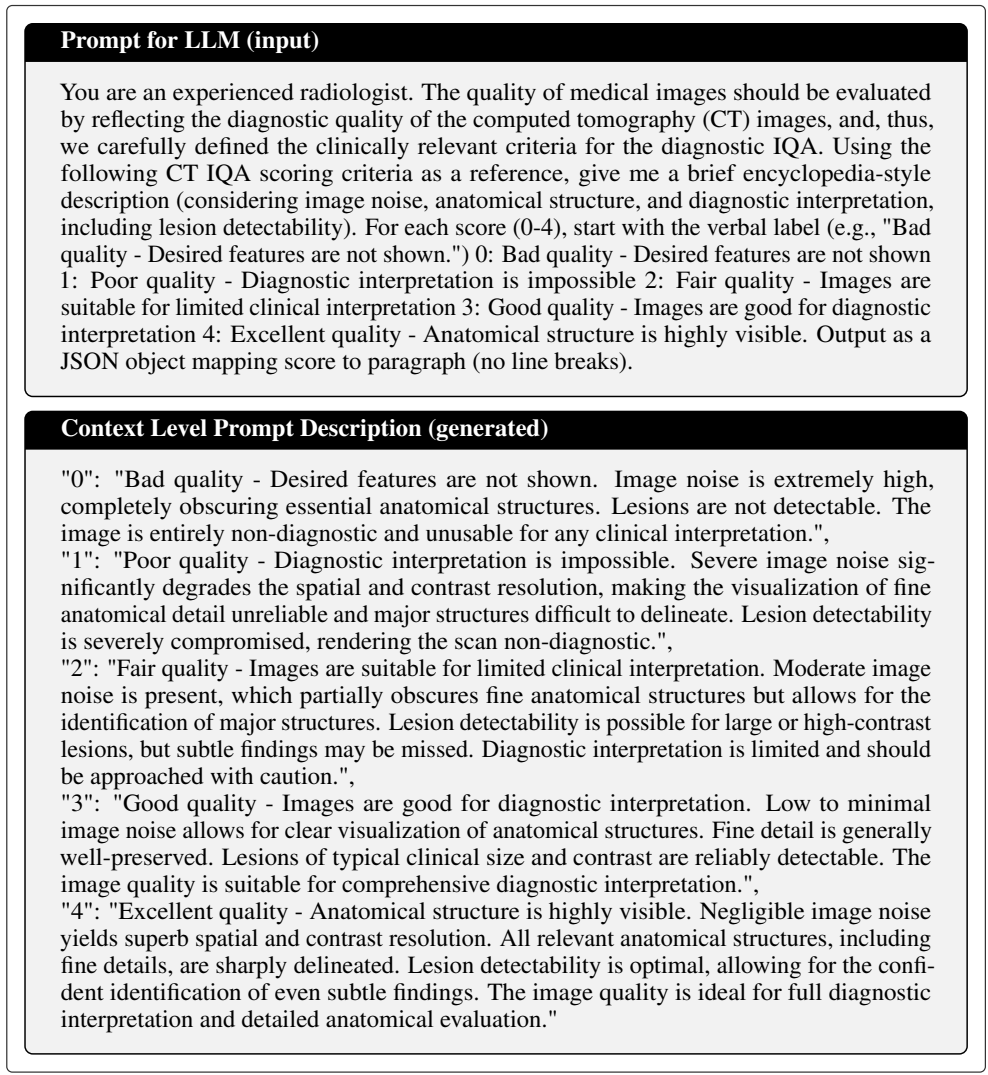

\centering

\resizebox{0.95\linewidth}{!}{

\tcbset{
  colback=white,
  colframe=black!80,
  boxrule=0.5pt,
  arc=2pt,
  left=4pt,
  right=4pt,
  top=4pt,
  bottom=4pt,
}

\begin{tcolorbox}[width=\linewidth]

  \begin{tcolorbox}[colback=gray!10, colframe=black, title=\textbf{Prompt for LLM (input)}, boxrule=0.6pt]
 You are an experienced radiologist. The quality of medical images should be evaluated by reflecting the diagnostic quality of the computed tomography (CT) images, and, thus, we carefully defined the clinically relevant criteria for the diagnostic IQA. Using the following CT IQA scoring criteria as a reference, give me a brief encyclopedia-style description (considering image noise, anatomical structure, and diagnostic interpretation, including lesion detectability). For each score (0-4), start with the verbal label (e.g., "Bad quality - Desired features are not shown.") 
 0: Bad quality - Desired features are not shown 
 1: Poor quality - Diagnostic interpretation is impossible 
 2: Fair quality - Images are suitable for limited clinical interpretation 
 3: Good quality - Images are good for diagnostic interpretation 
 4: Excellent quality - Anatomical structure is highly visible. 
 Output as a JSON object mapping score to paragraph (no line breaks). 
  \end{tcolorbox}
  
  \begin{tcolorbox}[colback=gray!10, colframe=black, title=\textbf{Context Level Prompt Description (generated)}, boxrule=0.6pt, breakable, listing only,
    listing options={
      basicstyle=\ttfamily\footnotesize,
      breaklines=true,
      columns=fullflexible,
      showstringspaces=false
    }]
{
  "0": "Bad quality - Desired features are not shown. Image noise is extremely high, completely obscuring essential anatomical structures. Lesions are not detectable. The image is entirely non-diagnostic and unusable for any clinical interpretation.", 

  "1": "Poor quality - Diagnostic interpretation is impossible. Severe image noise significantly degrades the spatial and contrast resolution, making the visualization of fine anatomical detail unreliable and major structures difficult to delineate. Lesion detectability is severely compromised, rendering the scan non-diagnostic.", 

  "2": "Fair quality - Images are suitable for limited clinical interpretation. Moderate image noise is present, which partially obscures fine anatomical structures but allows for the identification of major structures. Lesion detectability is possible for large or high-contrast lesions, but subtle findings may be missed. Diagnostic interpretation is limited and should be approached with caution.", 

  "3": "Good quality - Images are good for diagnostic interpretation. Low to minimal image noise allows for clear visualization of anatomical structures. Fine detail is generally well-preserved. Lesions of typical clinical size and contrast are reliably detectable. The image quality is suitable for comprehensive diagnostic interpretation.", 

  "4": "Excellent quality - Anatomical structure is highly visible. Negligible image noise yields superb spatial and contrast resolution. All relevant anatomical structures, including fine details, are sharply delineated. Lesion detectability is optimal, allowing for the confident identification of even subtle findings. The image quality is ideal for full diagnostic interpretation and detailed anatomical evaluation." 
}
  \end{tcolorbox}
\end{tcolorbox}
}
\caption{Generated text prompt description from Gemini for context-aware CT image quality assessment, including diagnostic visibility, interpretability, and artifact severity. The JSON structure for scores 0-4 is also presented.}
\label{fig:dose_prompt}
\end{figure}

\subsection{Context Generation for IQA}

\subsubsection{Prompting LLMs}
To evaluate the influence of large-language-model–generated textual priors, we compared CAP-IQA using prompts from ChatGPT~\cite{openai2025chatgpt}, Gemini~\cite{google2025gemini25pro}, Copilot~\cite{github2025copilot}, and Claude~\cite{anthropic2025claude_sonnet4_5}. Each model was instructed to produce clinical descriptions of CT image quality for scores 0–4, emphasizing image noise, anatomical visibility, and diagnostic interpretability. Gemini generated concise, radiologist-style definitions with balanced detail on lesion detectability and structural clarity, yielding the highest correlation scores (see Fig.~\ref{fig:barplot_iqa_scores}). ChatGPT provided comparable results with slightly less domain precision, whereas Copilot and Claude produced longer or less focused descriptions, which contributed to marginally lower performance. These results highlight that prompt quality and domain alignment directly affect the discriminative power of the textual priors in CAP-IQA.

Fig.~\ref{fig:dose_prompt} illustrates an example of the text prompt designed for context-aware CT image quality assessment. The prompt instructs the Gemini model to generate diagnostic-level descriptions that reflect the perceptual and clinical aspects of image quality, including noise, anatomical visibility, interpretability, and artifact severity. The model is guided by predefined scoring criteria ranging from 0 (bad quality) to 4 (excellent quality). Each score corresponds to a concise paragraph describing diagnostic relevance, structured in a JSON format. The example shown highlights the description for score 0, describing severely degraded image quality where anatomical structures and lesions are indistinguishable due to excessive noise or artifacts.

\subsubsection{Prompt Engineering}
\begin{figure}
    \centering
    \includegraphics[width=0.9\linewidth]{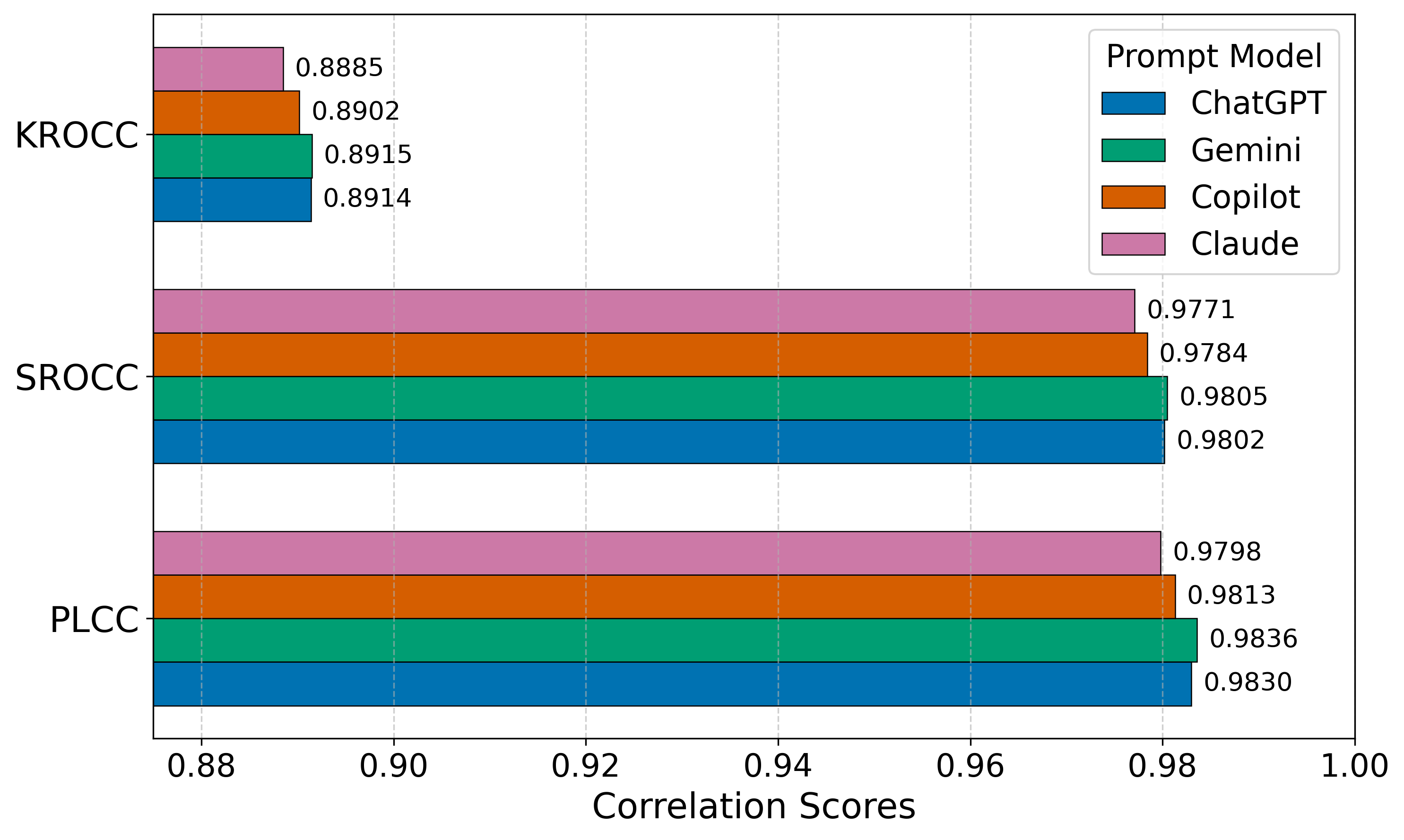}
    \caption{Grouped bar plot comparing the correlation metrics of CAP-IQA using different prompt generation models. Among all variants, Gemini-based prompts achieve the highest correlations, showing the influence of prompt quality on model performance.}
    \label{fig:barplot_iqa_scores}
\end{figure}

\begin{figure}
    \centering
    \includegraphics[width=0.9\linewidth]{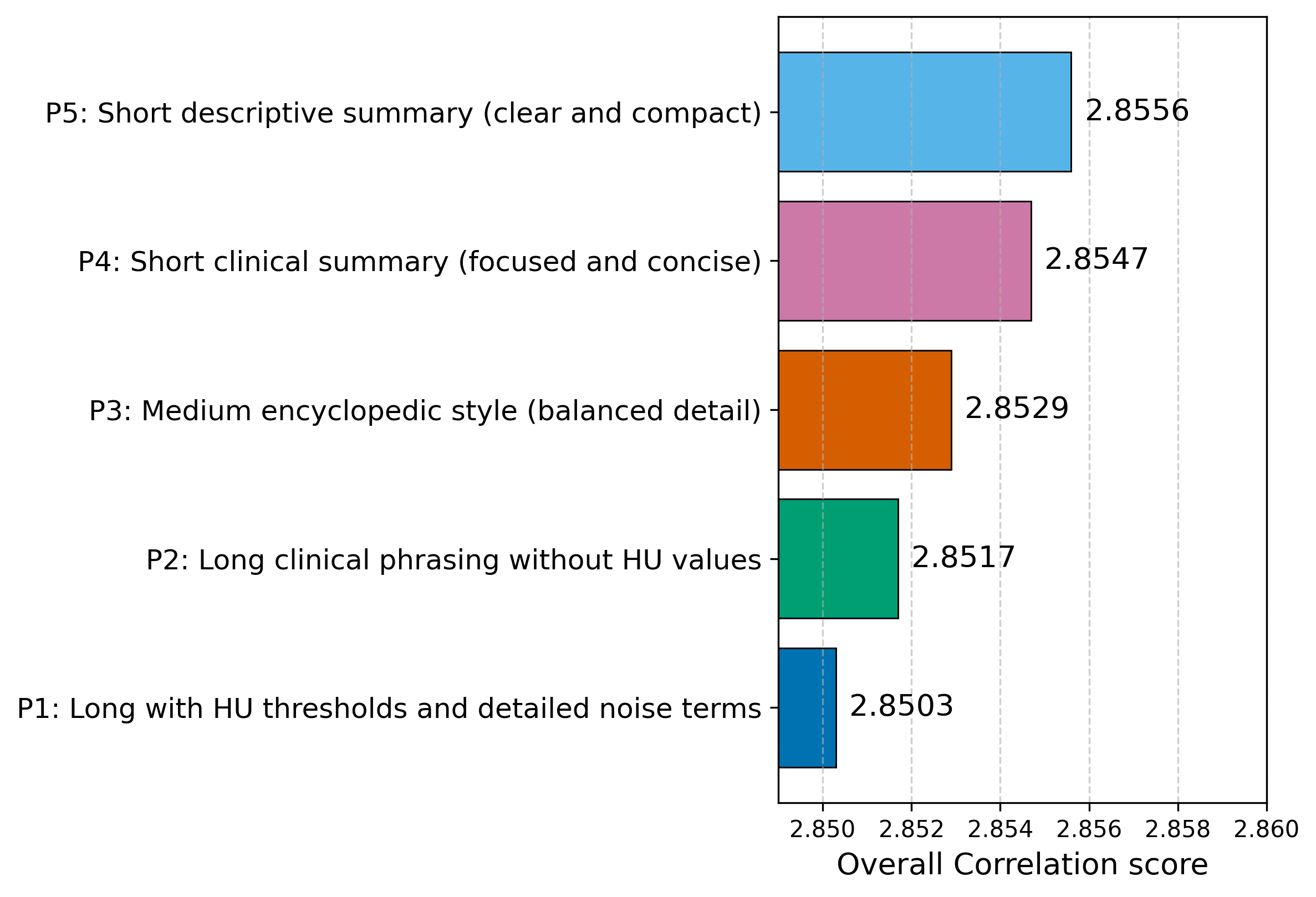}
    \caption{Effect of prompt variation on CAP-IQA performance. Each bar represents the overall correlation score for five Gemini-generated prompts with different styles and lengths. Shorter and more focused prompts achieved slightly higher correlations, highlighting the impact of prompt conciseness and clarity on IQA accuracy.}
    \label{fig:prompt_eng}
\end{figure}

For prompt engineering, we used the Gemini model, identified as the best-performing LLM in Fig.~\ref{fig:barplot_iqa_scores} for our CAP-IQA framework. We designed five Gemini-generated radiology-style prompts (P1--P5) that varied in wording length and level of clinical detail (Fig.~\ref{fig:prompt_eng}). These ranged from long, quantitative versions containing Hounsfield Unit (HU) thresholds and detailed radiology terminology to shorter, clinically phrased summaries. While all prompts followed the same diagnostic IQA criteria (0–4 scale), overly descriptive or numerically heavy prompts introduced redundancy and slightly reduced alignment between text and visual features. In contrast, concise and clinically coherent prompts achieved higher correlation consistency ($r$ = 0.9836, $\rho$ = 0.9805, $\tau$ = 0.8915) compared to longer ones ($r$ = 0.9810, $\rho$ = 0.9801, $\tau$ = 0.8892), reflecting a performance gap of about 0.2–0.3\%. These results suggest that shorter and concise prompts better capture diagnostic intent and contribute to more stable multimodal IQA performance.

\subsection{Real Clinical Evaluation}
\begin{figure}
\centering
\resizebox{0.85\linewidth}{!}{
\begin{tabular}{c c c}

\includegraphics[width=0.33\linewidth]{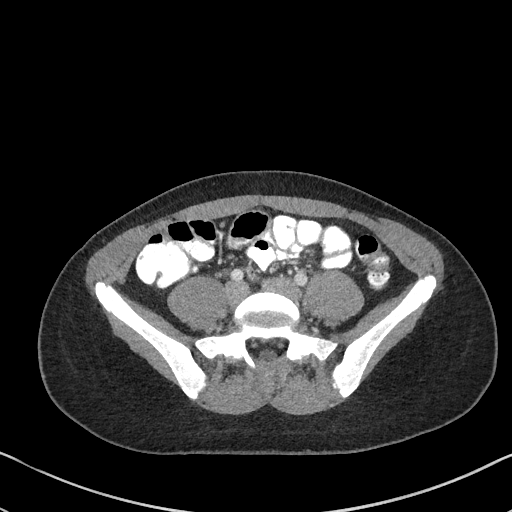}
&
\includegraphics[width=0.33\linewidth]{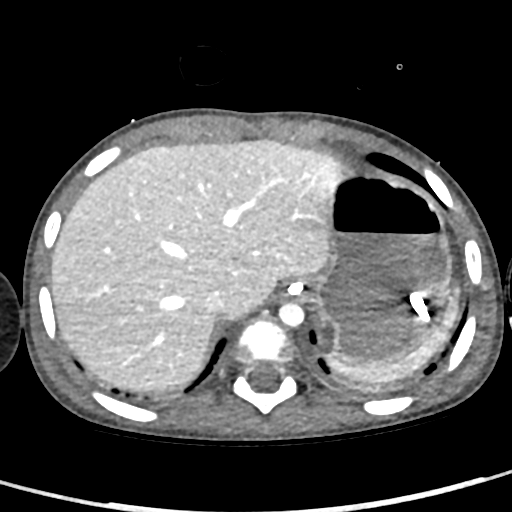}
&
\includegraphics[width=0.33\linewidth, trim={0in 0in 0in 0in}, clip]{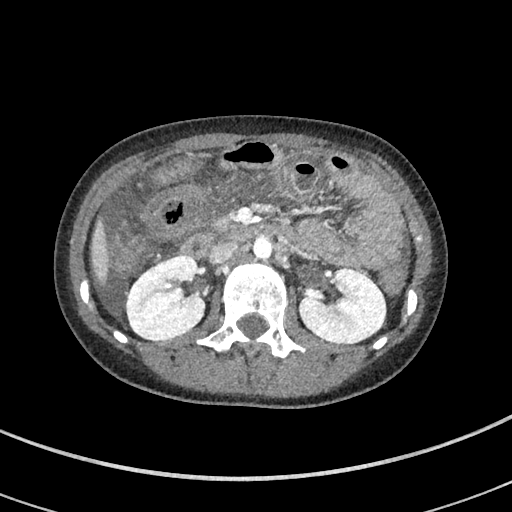}\\
{\large 3.95} & {\large 3.71} & {\large 3.94}
\smallskip\\
\includegraphics[width=0.33\linewidth]{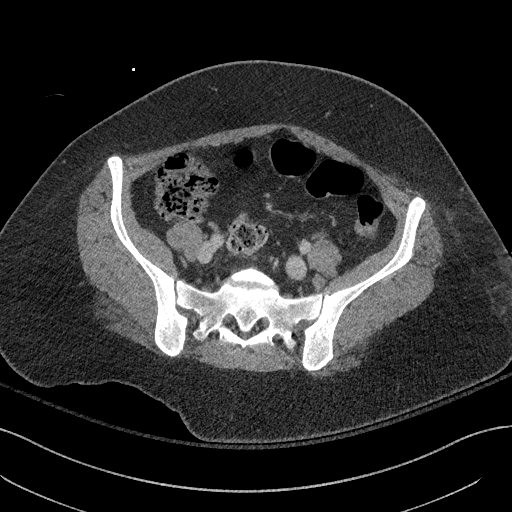}
&
\includegraphics[width=0.33\linewidth, trim={0in 0in 0in 0in}, clip]{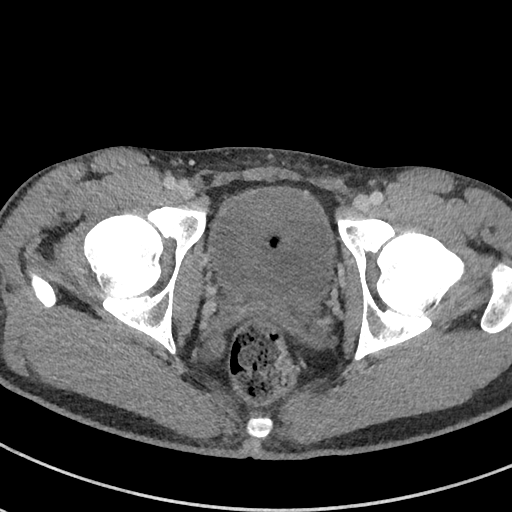}
&
\includegraphics[width=0.33\linewidth, trim={0in 0in 0in 0in}, clip]{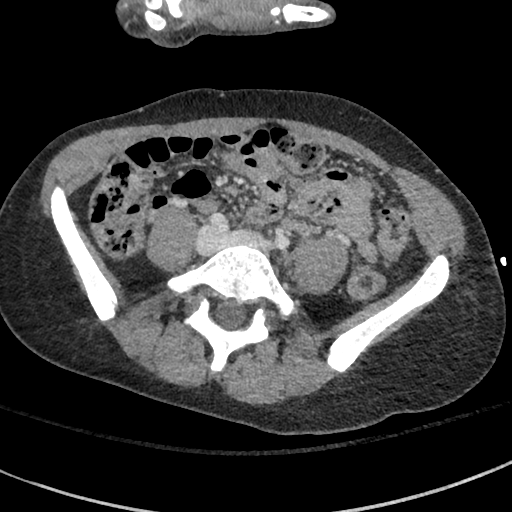}\\
{\large 3.46} & {\large 3.79} & {\large 3.55}
\end{tabular}
}

\caption{Visualization of sample CT image slices from the in-house pediatric scan dataset and corresponding predicted scores using CAP-IQA.[Window Width: 400 and Window Level: 50]}
\label{fig:ped_vis}
\end{figure}

\begin{figure}
    \centering
    \includegraphics[width=0.9\linewidth]{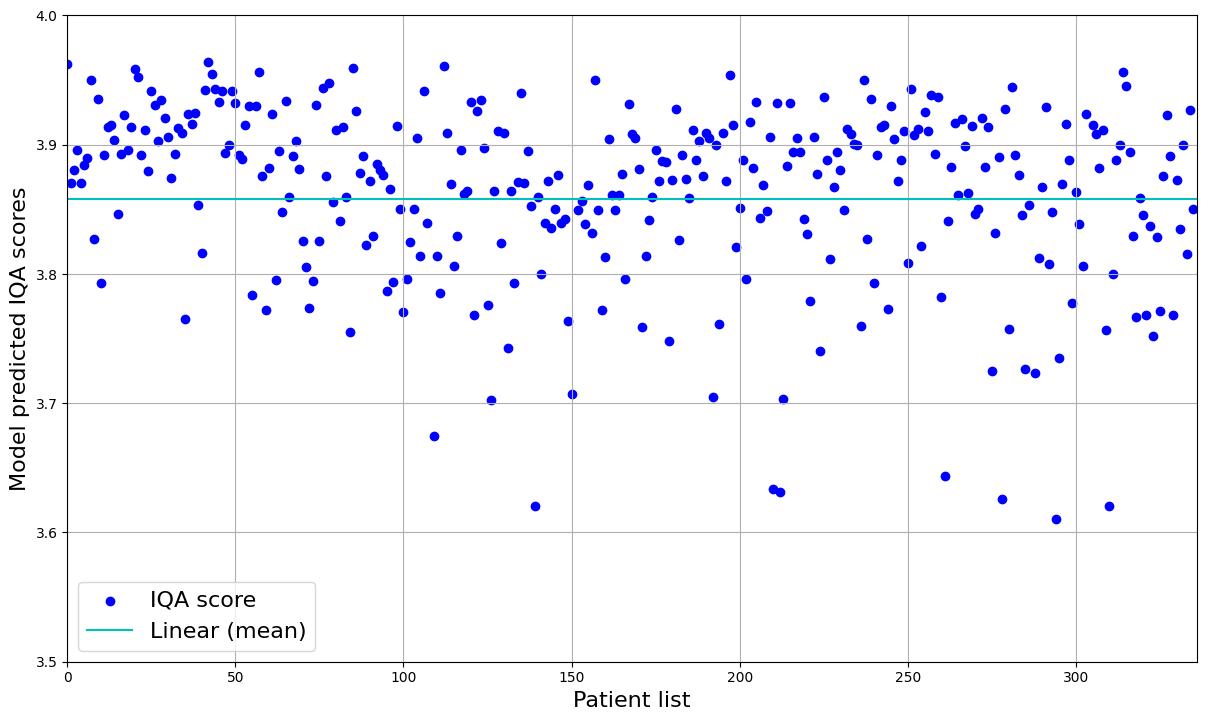}
    \caption{Distribution of predicted IQA scores across 336 pediatric CT exams, showing consistent and high diagnostic quality.}
    \label{fig:peds-iqa}
\end{figure}

We also evaluated the proposed CAP-IQA model on an in-house dataset. Fig.~\ref{fig:ped_vis} displays sample images from the dataset. We retrospectively collected real clinical CT images from the University of Kentucky Medical Center with approval from the Institutional Review Board. A total of 91,514 image slices from 336 pediatric patients aged 2-12 years were used for testing the generalizability of CAP-IQA. Since these images have already been used for clinical diagnosis, they are expected to be of high quality (>3, as per the IQA scoring criteria in Table~\ref{tab:score-criteria}). Slice-wise IQA scores predicted by the model were averaged to obtain the overall score for each of the CT scans. The average IQA score across the 336 pediatric scans is 3.8582, with a correlation of variation of 2.1447. As seen in Fig.~\ref{fig:peds-iqa}, the predicted scores are tightly grouped around the mean, with scores above the diagnostic-quality threshold of 3. The relatively high predicted scores align with expert radiologists' assessments, indicating that the scans retain good diagnostic quality. Overall, the results demonstrate that CAP-IQA performs reliably in real-world clinical settings, and the model consistently and accurately assesses quality across diverse patient data.

\subsection{Model Efficiency}
In terms of computational efficiency, CAP-IQA achieves an inference time of 52 ms per CT slice and a peak memory consumption of 358.5 MB, measured on an Intel Xeon W7-2475X (2.6 GHz) workstation with an NVIDIA A4000 GPU (16 GB). While the reported runtimes of LDCTIQAC 2023~\cite{lee2025low} models vary depending on implementation, most methods typically process a slice in 70–400 ms and use more than 500 MB of memory. In contrast, CAP-IQA demonstrates a faster and more lightweight design without sacrificing accuracy.

\section{Conclusion}
We have presented CAP-IQA, a context-aware prompt-guided framework that combines medical text priors with image-specific context prompts to assess CT image quality in a clinically meaningful way. CAP-IQA effectively distinguishes true image degradations, resulting in more reliable and interpretable predictions. Comprehensive evaluations on LDCTIQA reveal the superiority of our CAP-IQA by achieving the highest correlations with radiologist scores, over the top challenge algorithms and recent foundation-based models. Additionally, validation on real clinical CT images of pediatric patients demonstrates the true generalizability of CAP-IQA. Despite its effectiveness, the current framework is limited to 2D slice-level analysis and relies on pre-defined text prompts generated by large language models. This design may overlook 3D contextual dependencies and prompt variations that could affect interpretability in future unseen domains. In our future work, we plan to extend CAP-IQA to 3D volumetric IQA and incorporate more adaptive prompt tuning to better align with different scanning protocols and patient populations.

\bibliographystyle{agsm}
\bibliography{references}

\end{document}